Part of Speech and Universal Dependency effects on English Arabic Machine Translation

Authors: Ofek Rafaeli, Dr. Omri Abend, Leshem Choshen, dr. Dmitry Nikolaev

## Abstract

In this research paper, I will elaborate on a method to evaluate machine translation models based on their performance on underlying syntactical phenomena between English and Arabic languages. This method is especially important as such "neural" and "machine learning" are hard to fine-tune and change. Thus, finding a way to evaluate them easily and diversely would greatly help the task of bettering them.

## Introduction

Translation is the communication of the meaning of a source language text by means of an equivalent target language text. Machine translation is the act of translation done by machines (without the immediate intervention of human hands in the process). Machine translation has come a long way from the Georgetown–IBM experiment (1954)[1] through Automatic Language Processing Advisory Committee (1964), MOSES, the open -source statistical MT engine (2007)[2] and most recently and notably google translate, a service by Google that supports 109 languages (at various translation levels) and claims 100 billion words translated daily[3]. Both Human and Machine translations face numerous challenges: lexical ("what is the right word in the target language?"), semantical ("what is the correct use of a word?"), pragmatical ("what is the correct way to pass underlying meanings in a text?") and syntactical ("what is the correct structure that reflects the correct meaning in the target language?").

As stated above, machine translation has had a long history. In it, the translation process has changed approaches ranging from rule and transfer based methods, through inter-lingual, dictionary and statistical approaches and finally the current state of the art that is based on machine leaning, AI and neural networks.it is important to note on these last ones that they are highly reliant on the data they are "fed". This is very commonly discussed in machine learning literature as "Bias" towards such data. Bettering machine translation models in turn opens endless functionalities: from governmental, touristic, diplomatic actions and trading – easier translation from English to Arabic and vice versa are a gateway to a better future.

---

[1] Hutchins, W. John. "The Georgetown -IBM experiment demonstrated in January 1954." *Conference of the Association for Machine Translation in the Americas*. Springer, Berlin, Heidelberg, 2004.

[2] Koehn, Philipp, et al. "Moses: Open source toolkit for statistical machine translation." *Proceedings of the 45th annual meeting of the ACL on interactive poster and demonstration sessions*. Association for Computational Linguistics, 2007.

[3] https://www.blog.google/products/translate/ten -years -of -google -translate/

## Open challenges in machine translation

Koehn, Knowles (2017) identify challenges for machine translation:

1. Machine translation systems have lower quality out of domain (of which they were trained upon), to the point that they completely sacrifice adequacy for the sake of fluency. That is to say, that if a machine translation was trained upon a sports domain, it would act poorly on a weather domain.
2. Machine translation systems have a steeper learning curve with respect to the amount of training data (ie – they learn 'faster', perform better with more data), resulting in worse quality in low-resource settings (perform badly given small amounts of data), but better performance in higher source settings. That is to say, they have a 'bias' towards data-rich environments.
3. Machine translation systems have lower translation quality on very long sentences. This is especially interesting too point out in reagrds to Sulem, Abend, Rappoport's. "Bleu is not suitable for the evaluation of text simplification." arXiv preprint arXiv:1810.05995 (2018), as BLEU is commonly wide used for machine translation evaluation. Hence this point in particular might be debatable. I will elaborate on the mechanics of BLEU score later on in this paper.
4. The attention model for ' Machine translation systems ' does not always fulfill the role of a word alignment model, but may in fact dramatically diverge. This is a very important point for the purpose of this project. I will elaborate below regarding syntactic divergences that are closely correlated to this in Arabic in particular.
5. Beam search decoding strategies only improve translation quality and accuracy for narrow beams. It deteriorates when exposed to a larger search space. This point is important for the Verb and preposition distance phenomena when translating from Arabic to English, and I will delve deeper into this later on.

## Syntactic pattern arise

Hence, syntactic patterns in which languages differ are often used to describe work on cross-lingual transfer. Two famous examples of this are: (1) word order in sentences, such as Subject-Object-Verb (SVO) vs SOV, and (2) noun -adjective ordering ("the blue table" vs השולחן הכחול).

In this project, I aimed to suggest a method for automatic extraction of challenge sets for English -Arabic translation. The challenge sets are based off syntax phenomena and syntactic divergences in English and Arabic. Determining the phenomena in the challenge set was based on two key elements:

1. Data driven: manually tagging a Parallel Universal Dependency (henceforth PUD) corpus, I extracted common syntactic divergences that are assumed to be hard to translate.
2. Theoretical Arabic syntax phenomena: checking well -documented Arabic syntactic phenomena that do not have a fixed English parallel.

Following this, I manually extracted a couple of sentences of each suggested phenomena, showing that google translate fails to translate them accordingly. Finally, I aimed to extract (in an automatic fashion) such challenge sets from the PUD corpus using and evaluated the overall translation endeavor.

## Hebrew motivational Example:

Hebrew-English translation of sentiment verbs give an easy example of what we are looking for. Let us look at the following example:

**Hebrew:** דני מוצא חן בעיני דנה

**Literal English translation:** Danny finds beauty in dana's eyes.

**Golden English translation:** Dana Likes Dani

**Google translate output:** Danny likes Dana

Note that google translate switched the subject and the object of the liking: the Hebrew and the correct English translation both state that 'Danna' is the person doing the "liking". However, google translate determines that Danny is the one doing the "liking", and Danna is the recipient of this affection.

One might hypothesize that the reason for this mistake can be found in the syntactical position that "Danny" and "Dana" hold in both languages:

In Hebrew, דני holds the subject position. this can be proven as the verb is in the masculine form. Thus, even though semantically, this sentence points to Danna is the 'agent' in this sentence, she is only a passive person syntactically, occupying the object position.

Thus, one could bring up the notion that subject-object alternation might be a possible challenge for Hebrew - English translation. furthermore, one might want to check if other Hebrew sentiment verbs (such as "בא לי", "מחבב את", "סולד מ" and so fourth) present a similar fashion in their divergences.

Having this example in mind, what I aimed to do was find such common divergences in English and Arabic sentences, then "capitalize" on this knowledge to create a rule for extracting "challenge sets". These in turn would be used as a unit test – any translation machine that would assume to be good, must be able to translate such challenge sets correctly.

# Methodology

## Corpus

Universal Dependencies (henceforth UD[4]) is a framework for consistent annotation of grammar (parts of speech, morphological features, and syntactic dependencies) across different human languages. The Parallel Universal Dependency corpus (henceforth PUD) contains 1000 parallel in English and Arabic sentences.

The sentence alignment is one to one but occasionally a sentence -level segment actually consists of two real sentences. The sentences are taken from the news domain and from Wikipedia. There are usually only a few sentences from each document, selected randomly, not necessarily adjacent. 750 sentences are originally English. The remaining 250 sentences are originally from any of: German, French, Italian, or Spanish, and they were translated to other languages via English.

Professional translators translated the sentences from English to Arabic. Then the sentences were annotated morphologically and syntactically by Google according to Google universal annotation guidelines; finally, members of the UD community converted the sentences to UD v2 guidelines5.

## History Of Universal Dependencies[6]

The Stanford dependencies (which UD is based upon) were originally developed in 2005 as a backend to the Stanford parser to help in Recognizing Textual Entailment systems (such as: "I took cottage cheese out of the fridge" entailing "I took cheese out of the fridge"). The corpus eventually emerged as the de facto standard for dependency analysis of English. since then it has adapted to a number of different languages. The Google universal tag set grew out of the cross-linguistic error analysis. The corpus was then initially used for unsupervised part-of-speech tagging and has since been adopted as a widely used standard for mapping diverse tagsets to a common standard. Note this is the main point of this following project.

[4] https://universaldependencies.org/
[5] https://universaldependencies.org/treebanks/en_pud/index.html
[6] https://universaldependencies.org/introduction.html#history

The first attempt to combine Stanford dependencies and Google universal tags into a universal annotation scheme was the Universal Dependency Treebank (UDT) project, which released treebanks for 6 languages (2013) and 11 languages (2014). The second version of HamleDT provided Stanford/Google annotation for 30 languages in 2014. This was followed by the development of universal Stanford dependencies (USD). The new Universal Dependencies (again, Henceforth – UD) is the result of merging all these initiatives into a single coherent framework, based on USD.

The first version of the new tagging guidelines, (released October 2014), introduced an extended universal part-of-speech (POS) tag set. This set makes some distinctions that were missing in the original proposal, but were perceived to be of importance by many researchers, and clarifies the definition of categories. As a result of this work, universal POS categories have substantive definitions and are not necessarily just equivalence classes of categories in underlying language-particular treebanks. Hence, work to convert to UD POS tags often requires context-sensitive rules, or some hand correction (rather then just relaying on a machine parser). The UD morphological features aim to provide a stripped down basic set of features which are most crucial for analysis and are widespread across different languages. The dependency representation of UD evolves out of SD, which itself follows ideas of grammatical relations-focused description that can be found in many linguistic frameworks. That is, it is centrally organized around notions of subject, object, clausal complement, noun determiner, noun modifier, etc. The goal of the new universal version was to add or refine relations to better accommodate the grammatical structures of typologically different languages and to clean up some of the quirkier and more English-specific features of the original version. Hence, the new taxonomy has *Less* relations than the original SD.

## BLEU score – in depth

BLEU (bilingual evaluation understudy) is an algorithm for evaluation of quality of a text which is the result of machine-translation from one language to another language. Here, we define Quality as the correspondence between the machine's output and an output generated manually by a human. the closer the machine translation is to a professional human translation (a result of a human translator), the better it is. this is the central idea behind BLEU. BLEU was one of the first metrics to claim a high correlation with human judgements of quality, and remains one of the most popular automated and inexpensive metrics.

BLEU Scores are calculated for individual translated segments. That is to say that it usually deals with individual sentences. It does so by comparing them (the machine translation of such sentences) with a set of good quality reference translations. Those scores are averaged over the whole corpus to reach an estimate of the translation's overall quality. Intelligibility or grammatical correctness are not taken into account. This means that BLEU has more to do with the "accuracy" of translation (in comparison with a golden standard) rather than with the "fluency" of such translations.

BLEU's output is always a number between 0 and 1. This value indicates how similar the candidate text is to the reference texts, with values closer to 1 representing more similar texts. It is interesting to note that few human translations will attain a score of 1, since this would indicate that the candidate is identical to one of the reference translations. For this reason, it is not necessary to attain a score of 1 for a translation to be "good". Because there are more opportunities to match, adding additional reference (more "golden standard" – correct sentences) translations will increase the BLEU score of a corpus.

## Manual Tagging

Sentence level alignment is too coarse grained to extract syntactic insights (as traditional syntax deconstructs the sentence into subdivisions). For a data driven hypothesis, a more fine -grained analysis was needed. Hence, I manually went through each of the parallel 1000 sentences in English and Arabic, matching every content word[7] in both languages. Note I am not a native Arabic, but I have studied written Arabic (الفصحى) at an intermediary university level, and have achieved R -1+ level[8]. Additionally, note that this task did not require me to actively translate between English and Arabic, but only to match content words. Thus I hope to assume that I did in fact do this with a high degree of accuracy.

Finally, among the 1000 sentences, 2 sentences were not fully tagged:

- Sentence 454:
  English: *It contains a tiny lagoon, which has all but dried up.*
  Arabic: *تضم هوراً صغيراً لا يزال مملوءاً بالماء.*
  the Arabic sentence refers to a lagoon that is still filled with water
- Sentenc 491:
  English: *Cape Town's local government is the City of Cape Town, which is a metropolitan municipality.*

[7] "Content word" definition and tagging instructions: https://docs.google.com/document/d/12QnVRyTMPDhSw9QT_aA4aS_qFuamdPBONbKwaFv86-k/edit?ts=5dbc8936

[8] https://www.govtilr.org/Skills/ILRscale4.htm

Arabic (gloss form): الحكومة المحلية لكيب تاون هي بلدية كيب تاون, وهي بلدية شاملة للمنطقة الحضرية

here, the Arabic word بلدية does not correspond to *municipality*, but rather to a physical entity (similar to Hebrew "עירייה"). In that sense, it takes a physical being that turns to be the subject of the sentence, changing the meaning from the English original.

## Mid -level results

Once I finished tagging the sentences, I delve into the data, extracting the following high -level summaries:

*Table1 : POS correlation matrix percentages*

| X | VERB | SYM | SCONJ | PUNCT | PROPN | PRON | PART | None | NUM | NOUN | DET | CCONJ | AUX | ADV | ADP | ADJ | en |
|---|---|---|---|---|---|---|---|---|---|---|---|---|---|---|---|---|---|
| 0 | 2.2 | 0 | 0 | 0 | 1.4 | 0.2 | 0.5 | 2.5 | 0.1 | 16.8 | 1.8 | 0 | 0.1 | 1 | 1.7 | 71.7 | ADJ |
| 0 | 9.6 | 0 | 0 | 0 | 0.2 | 0.4 | 0.2 | 5.5 | 0 | 14.5 | 0.4 | 0 | 0.4 | 1.7 | 62.9 | 4.2 | ADP |
| 0.2 | 5.2 | 0 | 0 | 0.2 | 0 | 0.7 | 8.9 | 5.6 | 0 | 16.3 | 0.4 | 0.4 | 0.7 | 39.9 | 9.3 | 12.2 | ADV |
| 0 | 62.7 | 0 | 0.6 | 0 | 0 | 2.3 | 9.3 | 4.7 | 0 | 3.9 | 0.2 | 0 | 13.8 | 0.4 | 1.7 | 0.4 | AUX |
| 0 | 0 | 0 | 0 | 0 | 0.9 | 0 | 24.8 | 0 | 0 | 2.8 | 0.9 | 31.2 | 0 | 0.9 | 37.6 | 0.9 | CCONJ |
| 0 | 0.9 | 0 | 0 | 0 | 0.9 | 35 | 4.2 | 1.4 | 0 | 19.2 | 24.8 | 0 | 0 | 3.3 | 4.7 | 5.6 | DET |
| 0 | 1 | 0.1 | 0 | 0 | 3.6 | 0.1 | 0 | 1.9 | 0.2 | 87.2 | 0.3 | 0 | 0.1 | 0.6 | 0.3 | 4.8 | NOUN |
| 0 | 0 | 0 | 0 | 0 | 0 | 0.7 | 0 | 1.6 | 69 | 11.8 | 6.2 | 0 | 0 | 0 | 0 | 10.8 | NUM |
| 0 | 13.9 | 0 | 0.1 | 0 | 2.8 | 15.2 | 1.3 | 0 | 0.1 | 41.9 | 1.6 | 0.3 | 2.3 | 1.9 | 6.6 | 12 | None |
| 0 | 7.7 | 0 | 2.2 | 0 | 0 | 1.1 | 52.7 | 2.2 | 0 | 12.1 | 0 | 0 | 4.4 | 0 | 16.5 | 1.1 | PART |
| 0 | 17 | 0 | 0.3 | 0 | 0.1 | 70.9 | 0.3 | 2.8 | 0 | 3.3 | 0 | 0 | 1 | 3 | 1.2 | 0.1 | PRON |
| 0 | 0.1 | 0 | 0 | 0 | 83.1 | 0 | 0 | 0.3 | 0 | 8.7 | 0.1 | 0 | 0 | 0 | 0 | 7.8 | PROPN |
| 0 | 0 | 0 | 0 | 50 | 0 | 50 | 0 | 0 | 0 | 0 | 0 | 0 | 0 | 0 | 0 | 0 | PUNCT |
| 0 | 1.2 | 0 | 3.6 | 0 | 1.2 | 1.2 | 1.2 | 0 | 0 | 7.1 | 0 | 0 | 0 | 0 | 81 | 3.6 | SCONJ |
| 0 | 0 | 68 | 0 | 0 | 0 | 0 | 0 | 0 | 0 | 32 | 0 | 0 | 0 | 0 | 0 | 0 | SYM |
| 0 | 62.6 | 0 | 0 | 0 | 0.3 | 0.2 | 0.1 | 1.7 | 0 | 24.9 | 0.1 | 0 | 2.2 | 0.6 | 1.5 | 5.9 | VERB |
| 0 | 0 | 0 | 0 | 0 | 100 | 0 | 0 | 0 | 0 | 0 | 0 | 0 | 0 | 0 | 0 | 0 | X |

Every row in the table above refers to a Part of speech (POS) English, and every column refers to a POS in Arabic. The number that appears in row i, column j refers to the percentage of words that have the tag i in English, and the tag j in Arabic. Thus, the main diagonal (colored in blue and azure) refers to the percentages of matching tags (that are not the focus of this study).

Note that two different matrices were examined:

1. Count of the amount of appearances sharing these two tags in the two languages (see example below).
2. The percentage out of the English words with this tag to have this matching tag in Arabic.

Orange colored cells refer to phenomena that both have relatively high percent (over 8%), and non -negligible counts (over 50 appearances).

*Table2 : UD correlation matrix counts*

| xcomp | parataxis | obl | obj | nummod | nsubj | nmod | flat | fixed | conj | compound | ccomp | appos | amod | advmod | advcl | acl | en |
|---|---|---|---|---|---|---|---|---|---|---|---|---|---|---|---|---|---|
| 2 | 0 | 6 | 0 | 0 | 4 | 32 | 0 | 0 | 3 | 0 | 13 | 0 | 10 | 1 | 11 | 117 | acl |
| 20 | 0 | 52 | 0 | 0 | 1 | 11 | 1 | 0 | 5 | 0 | 2 | 0 | 0 | 1 | 44 | 6 | advcl |
| 1 | 0 | 43 | 5 | 0 | 3 | 27 | 0 | 0 | 1 | 21 | 0 | 0 | 12 | 181 | 1 | 0 | advmod |
| 0 | 0 | 40 | 2 | 1 | 5 | 106 | 3 | 1 | 0 | 1 | 14 | 2 | 766 | 7 | 0 | 8 | amod |
| 0 | 0 | 1 | 0 | 0 | 1 | 12 | 0 | 0 | 1 | 0 | 0 | 56 | 1 | 0 | 3 | 1 | appos |
| 11 | 0 | 4 | 0 | 0 | 1 | 1 | 0 | 0 | 1 | 0 | 50 | 0 | 0 | 0 | 1 | 2 | ccomp |
| 0 | 0 | 5 | 1 | 11 | 3 | 277 | 47 | 0 | 3 | 3 | 1 | 5 | 184 | 0 | 0 | 0 | compound |
| 0 | 0 | 0 | 0 | 0 | 0 | 8 | 1 | 0 | 318 | 0 | 0 | 0 | 1 | 0 | 23 | 0 | conj |
| 0 | 0 | 1 | 0 | 0 | 0 | 0 | 1 | 0 | 0 | 0 | 0 | 1 | 1 | 0 | 0 | 0 | fixed |
| 0 | 0 | 0 | 0 | 0 | 0 | 7 | 123 | 0 | 0 | 0 | 0 | 36 | 11 | 0 | 0 | 0 | flat |
| 0 | 0 | 30 | 4 | 1 | 7 | 717 | 2 | 0 | 4 | 0 | 0 | 3 | 10 | 1 | 2 | 0 | nmod |
| 0 | 0 | 19 | 27 | 0 | 696 | 41 | 0 | 0 | 0 | 0 | 4 | 1 | 1 | 1 | 0 | 7 | nsubj |
| 0 | 0 | 14 | 0 | 81 | 0 | 3 | 0 | 0 | 0 | 0 | 1 | 0 | 15 | 1 | 0 | 0 | nummod |
| 0 | 0 | 94 | 279 | 0 | 24 | 153 | 0 | 0 | 0 | 0 | 0 | 0 | 2 | 1 | 2 | 4 | obj |
| 2 | 0 | 423 | 22 | 0 | 28 | 81 | 0 | 0 | 0 | 0 | 1 | 1 | 0 | 4 | 10 | 0 | obl |
| 0 | 5 | 0 | 0 | 0 | 0 | 0 | 0 | 0 | 6 | 0 | 21 | 0 | 0 | 0 | 6 | 1 | parataxis |
| 12 | 0 | 64 | 23 | 0 | 6 | 15 | 0 | 0 | 1 | 0 | 7 | 0 | 0 | 3 | 4 | 2 | xcomp |

All of the raw tables and the python script that was used to extract the correlation matrices are attached to this paper.

Following this data, I aggregated all sentences from the manually tagged corpus that contained one the of the following divergences:

obl -> nmod, amod -> nmod, Aux -> verb, obj -> nmod, Verb -> Noun, xcomp -> obl. Using the original (professionally translated) Arabic sentence as the golden standard, I used the BLEU score to check how well google translate performed, arriving at the following results:

| phenomena | BLEU score mean | Number of sentences |
|---|---|---|
| obl -> nmod | 0.70539172 | 81 |
| amod -> nmod | 0.693759802 | 113 |
| Aux -> verb | 0.702538114 | 129 |
| obj -> nmod | 0.702711357 | 144 |
| Verb -> Noun | 0.701787895 | 307 |
| xcomp -> obl | 0.703628831 | 74 |

Looking over the whole data set (all of the 1000) sentences the mean bleu score was: 0.7495161015312339. that is to say that this inquiry did not provid a sufficient challenge set: Google Translate's performance on these sentences was better than the average performance.

Using this link, any spectator can play with the data themselves. Note that for calculation I've used nltk translate bleu_score sentence_bleu for calculating the actual bleu score average usinga pandas dataframe framework (called challenge_set in the notebook). More on this package can be read here and here. Note this function uses a smoothing function to avoid the following edge case:

If there is no ngrams (consecutive target language words) overlap for any order of n-grams, BLEU (might) return the value 0. This is because the precision for the order of n-grams without overlap is 0, and the geometric mean in the final BLEU score computation multiplies the 0 with the precision of other n-grams. This results in 0 (independently of the precision of the other n-gram orders). This probably isn't a problem with the relatively high-results obtained above.

The default BLEU calculates a score for up to 4-grams (ie n=4) using uniform weights (this is called BLEU-4).checking the above sentences with n=5, and n=6 and n=3 did not alter the results greatly (using uniform weights).

During the work process, it has been suggested that BLEU isn't the most appropriate method for automatic checking of translation quality (see "Bleu is not suitable for the evaluation of text simplification." arXiv preprint arXiv: 1810.05995 (2018) ). To accompany for it's faults, it was suggested to add a check on the translated sentence along the lines of: "does it contain the expected syntactical formulation". For example, if the original (English) sentence contains an auxiliary verb, checking if the target sentence contains a verb. However, this approach lacks in two important aspects:

1. As can be seen from the statistics above, we do not expect these divergences to exist in 100% of the times. In fact, this is the surprising point of this study. Hence, doing this check contains the risk of penalizing "correct" sentences that do not contain the divergence.
2. This hypothetical check aims to see if the source word tagged with a certain tag corresponds to the target word containing the target tag. Unfortunately, doing so would either relay on manual checking of the target sentences corresponding words (similar to the manual tagging that I preformed, described above), or relay on another layer of machine learning, that matches every single word in every single sentence. This opens up a possibility of fine grained error that doesn't even have to do directly with the work in this project. As this would introduce irrelevant noise, it was decided not to contain this effort as a part of my paper, and leave this avenue of research for future endeavors.

## Noteworthy divergent constructions

### AUX -> VERB:

Auxiliary verb to actual verb. Looking at the data for any particular example that had more than 3 occurrences (of the same two word being aligned):

- English inflections of "be": "was", "were", "am" etc align with the arabic word كان ("was") and it's inflections (كانت ,يكون ,تكون etc)
- English: "can", "could", "have" lign with the arabic word يمكن ("will be possible\could")

Example, sentence 505:

English: The area controlled by the Bogd Khaan **was** approximately that of the former Outer Mongolia during the Qing period.

Arabic: **كانت** المنطقة التي حكمها بوغد خان هي تقريباً ما كان يدعى سابقاً بمنغوليا الخارجية خلال حكم سلالة تشينغ

Literal translation: "was the area that Bogd Khaan controlled it approximately what was known formerly as outer Mongolia during the Qing period".

English Sentence

## VERB -> NOUN:

looking at the set of two words most-aligned (4 times and above):

- English 'According' with Arabic 'وفقاً'
- English 'working' with  Arabic 'العمل' (literally: the work, or, the business)
- English 'cross' with Arabic 'عبور'

Example sentence 620:

English: **According** to a genetic study on autosomal data on Roma the source of Southasian Ancestry in Roma is North - West India

Arabic: **وفقاً** ل دراسة جينية على البيانات الصبغية الخاصة ب الغجر , ف إن الأصل الآسيوي الجنوبي عند الغجر يعود إلى شمال غرب الهند.

Literal translation: standing to study of genetics on special cell studies in Rome, indeed the source of Southasian Ancestry in Roma returns to north west india.

## amod -> nmod:

An adjectival modifier to a noun modifier. I could not find many same word-examples that repeat themselves. However, I do supply an example below that is representative of this nature:

Example, sentence 122:

English: The new iron guidelines mean **more donors** are needed.

English UD tree:

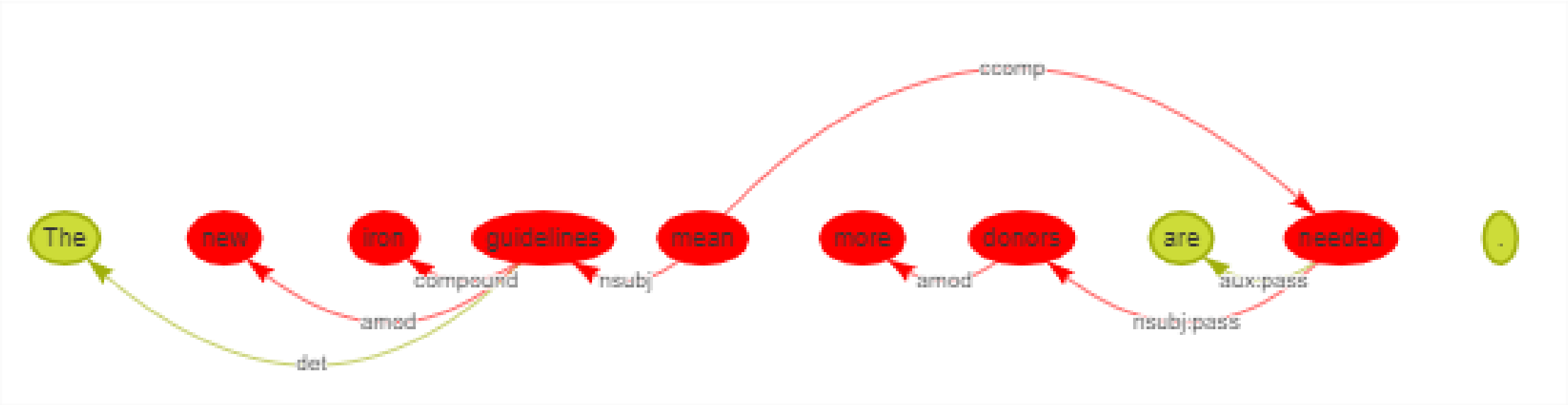

Arabic:  إرشادات الحديد الجديدة تعني أن هناك حاجةً ل**لمزيد** من **المتبرعين**.

Arabic UD tree:

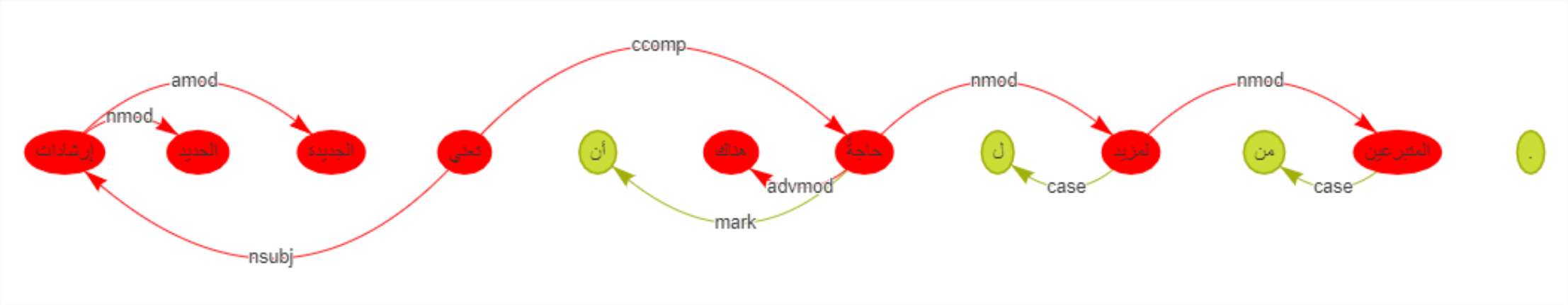

literal translation: the guidelines iron  the new mean that there's a need for more of the donors

(مزيد  - more,  متبرع  - donor, **المتبرعين**  - the donors)

## xcomp -> obl:

An open clausal complement to a nominal (noun, pronoun, noun phrase) functioning as a non-core (oblique) argument or adjunct. Again, I was not able to find repeating particular examples, however, following is an example of this divergence:

Example sentence 487:

English:  During the first century of development , Spanish dominance in the region **remained undisputed**

English UD tree:

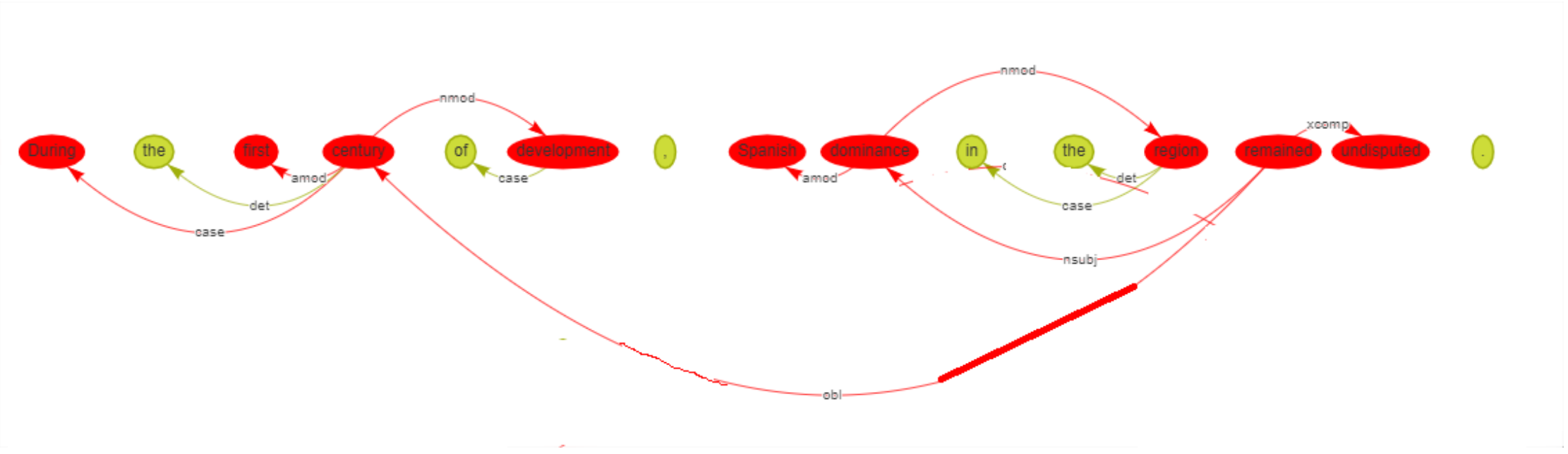

Arabic:منازع هيمنة الإسبان على المنطقة ب لا **ظلت** , خلال القرن الأول ل لتنمية .

Arabic UD tree:

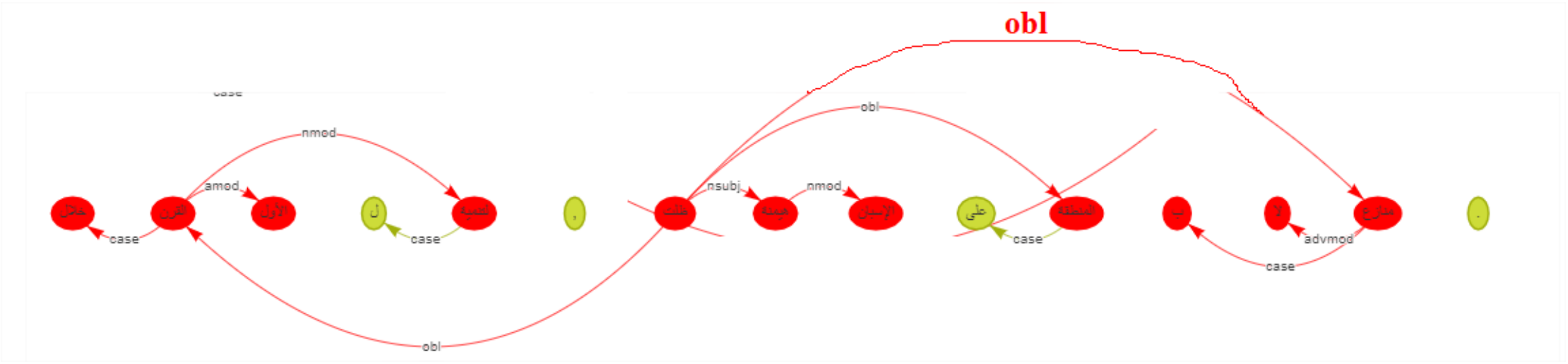

Literal translation: during the century the first for development, remained the dominance the spanish on the region not disturbed.

( ظلت - remained,  منازع - Contested, challenged, disturbed)

obj -> nmod:

an object's verb in English translated to to a noun modifier in Arabic

Example sentence 179:

English: They have one crack at redemption , **beating England**.

English UD sentence tree:

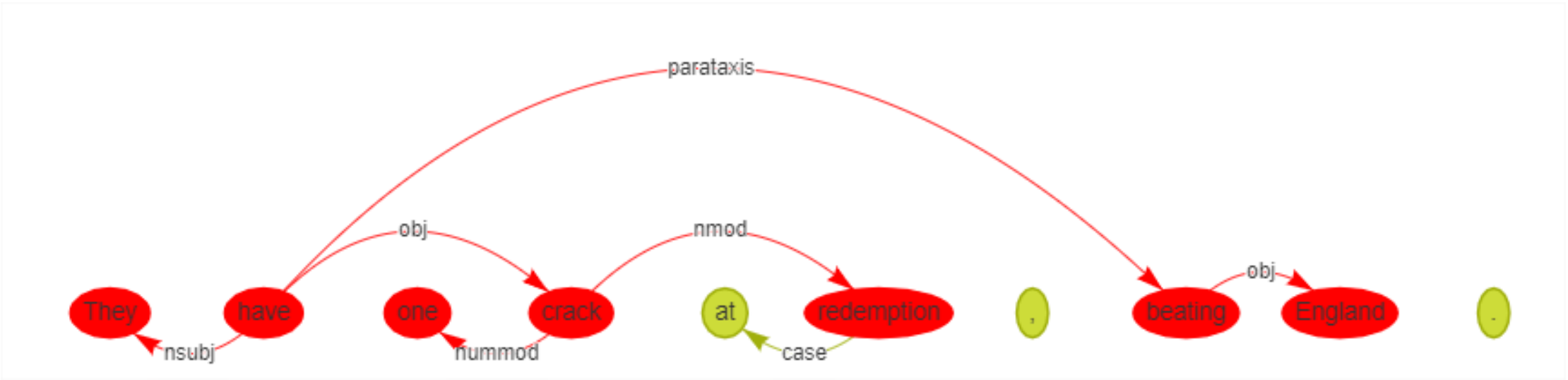

Arabic: أمام هم عائقٌ واحدٌ للخلاص , **إنجلترا إلى التغلب**

Arabic UD sentence tree:

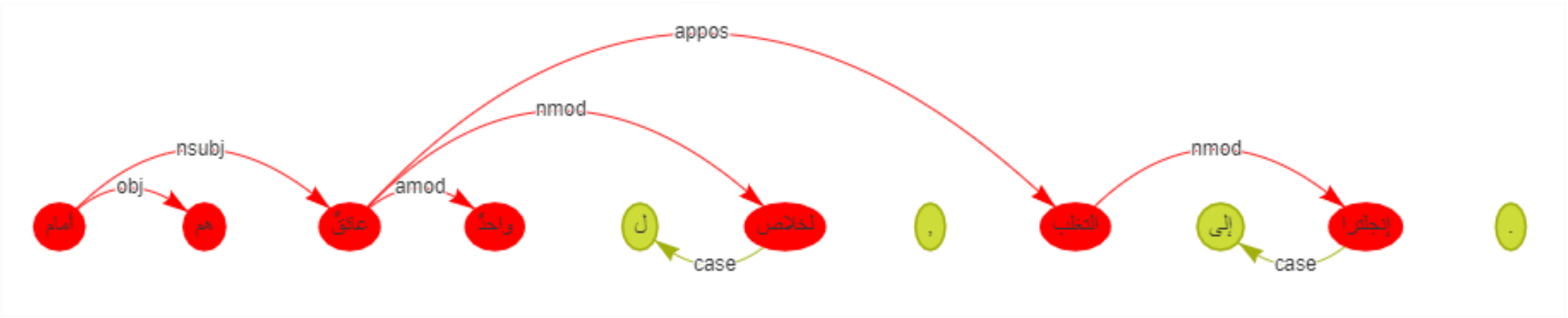

literal translation: in-front (of) them obstacle single for  salvation, the overcoming to England

( إنجلترا - England ,  التغلب - overcome,  overpower)

obl -> nmod :

- sentence : 37

english: Most people would find airport wi-fi more useful than being able to **send** emails on a **plane**

arabic: الرسائل **إرسال** قد يجد أغلب الناس أن " الواي - فاي " في المطارات أكثر فائدة من القدرة على **الطائرات** الإلكترونية داخل .

literal translation: may find most people that "the wi fi" in the air ports (is) more beneficial than the ability to send the letters the electronic inside the airplanes

( إرسال - send/sending , الطائرات - airplanes )

## Other ideas

During the manual tagging of the data, I aggregated a list of Arabic -unique syntactic phenomena (or: phenomena that do not appear in English) that did in fact appear in the set. I hypothesized that these might be hard to translate as they require a change in syntactic form, pro -drop or the like. These following phenomena were then checked manually for examples where a translation engine performed badly upon them:

### Passive Particle:

In Arabic, most verbs have a passive voice participle. This meaning is derived not from a syntactic change in the sentence ("I ate the apple" -> "the apple was eaten") but rather from a change in the verb structure ("اكلت تفاحة" -> "يؤكل تفاحة"). It should be noted that UD does include data on passive voice, however I could not find out a way to separate passive verbs and passive nouns (i.e. בינוני פעול) as sentence level passive voice.

Example:

Arabic Sentence: **ظفرت** الجائزة بتصفيق

Correct English sentence: The award was received with applause

Google translate: The award won applause

### Dual form:

In addition to singular and plural, Arabic has a dual form that refers specifically to two of a certain entity. No dual noun or verb information was found in UD.

Example:

Arabic Sentence: لنفسه اراد **الطفلين**

Correct English sentence: He wanted the two children for himself

Google translate: The two kids wanted for themselves

### Maf'ul mutlaq: (cognate accusative)

A verbal noun which comes after the verb, and it resembles the verb in form, or sometimes in meaning. It is used for either emphasis, stating a type or stating a number. As the root field for most words in the Arabic UD parser is not filled (it usually only appears on the verb, and not on the advmod, where it is needed), extracting these sentences proved difficult. Trying to automatically solve this, I searched for algorithms that extract roots. These proved ineffective, as they either stood around 75% accuracy, had a hard time with the suffix (such as ـاً ,which is the original purpose of this query), or were too slow to run on my computer fast enough (models were not published and only the training technique without data was published)[9]

Example:

English: she is closely linked
Arabic: ترتبط ارتباطاً وثيقاً
Literal translation: she is linked a strong link

### Verb and preposition distance:

This phenomenon is consistent with what is presented in -Choshen, Abend. "Automatically Extracting Challenge Sets for Non -local Phenomena Neural Machine Translation." arXiv preprint arXiv: 1909.06814 (2019), and hence was not delved into.
Example:

Arabic Sentence: **خرج** الجندي **على** القائد

Correct English sentence: The soldier attacked the commander

Google translate: The soldier came out on the commander

## Comments on the work process

While I do not hold that I am in any way fluent in spoken or written Arabic, I have however dedicated several years to learning the language. I've spent all 3 years of my high school education studying the language. During my 4 years of military service, I used to occasionally read Arabic newspapers as a leisure activity. During my university studies, I've spent a year studying Arabic as an advanced course (achieving R -1+ level). I do not delve here into my English knowledge, as a Linguistics undergraduate student is expected to have a control of the language that should suffice for this task.

[9] See:
https://www.sciencedirect.com/science/article/pii/S1319157815000166
https://www.researchgate.net/publication/221592138_Enhanced_Algorithm_for_Extracting_the_Root_of_Arabic_Words
https://ieeexplore.ieee.org/stamp/stamp.jsp?arnumber=8686325

All this is to say that replicating this study with more language pairs requires a certain level proficiency in these said languages. It is not my place to determine the criteria for this. However, as a future recommendation for any person that wishes to replicate or verify this kind of procedure, I would suggest at the very least being able to read a given sentence from a newspaper or Wikipedia article without having to consult the dictionary for more than half the words. Additionally, having studied the "formal" grammatical structures of a languages is extremely helpful when thinking about problematic divergences. This is especially the case if you have had the opportunity to study (As an adult) both languages you are dealing with (such as my personal case with the Israeli education system's teachings of Arabic and English).

## Summary

Machine translation is an up-and coming tool and research subject. Achieving better translation quality (and accuracy) is a difficult task. One approach to do this is to create a challenge set. That is, a set of sentences in a source language, with a corresponding golden translation sentences in a target language. Such a set must be "difficult" in the sense that any "reasonable" machine translation engine must perform well on translating the source language set to the target language. Different studies and methods for evaluation have been suggested; see BLEU score and -Sulem, Abend, and Rappoport. "Bleu is not suitable for the evaluation of text simplification". In this study I aimed to suggest an automatic way to extract challenging sentences for English - Arabic translation, based on syntactic differences in target and source language. Notice that this study has aimed not to extract a specific challenge set (that in turn would be used to evaluate a machine translation model), but to extract a rule that would make it easier for people to extract their very own challenge sets in an automatic fashion. This minor, yet important distinction has to do with the the spirit of the effort to better machine translation as a whole: human hands and time are an expensive resource, thus we would like to both diversify the inputs a machine translation engine receives, without requiring too much direct human effort going into it. Thus, this study has aimed to cross-linguistically better and facilitate the process of evaluating (and assuring) the quality of such translation models.

Sadly, it seems that the specific set of tags that rose from the manual tagging did not yield problematic results. That is to say, the google translation engine preformed relatively well on this set (at least compared to its own performance on a larger set of sentences). Lastly, I suggested more endeavors to find challenge sets automatically that have proven larger than the scope of this study. As future steps, I suggest either delving into the steps above, applying the same methods in this study to the reverse translation process (source language: Arabic, target language: English) or checking the same set of sentences according to a different matric: either semantic based ones, or checking that the suggested UD/POS tags appear in a target language.

I opened this paper with an Italian proverb, and I will finish it with a quote that is attributed to both the Hebrew Poet Hayim Nahman Bialik and the Canadian Poet Anne Michaels:

*"Reading a poem in translation is like kissing a woman through a veil" - Anne Michaels*

*"לקרוא שירה מתורגמת זה כמו להתנשק מבעד למטפחת" – חיים נחמן ביאליק*

## Appendix

- en -ar_pos_cm_percent.xlsx
- en -ar_pos_cm.xlsx
- en -ar_path_cm_no_subcats.xlsx
- compute_confusion_matrices.py
- challenge_set.xlsx